%% file: main.tex
\documentclass[10pt,twocolumn,letterpaper]{article}

\usepackage{iccv}              

\usepackage{booktabs}
\usepackage{multirow}
\usepackage[normalem]{ulem}
\useunder{\uline}{\ul}{}
\usepackage{graphicx}


\definecolor{iccvblue}{rgb}{0.21,0.49,0.74}
\usepackage[pagebackref,breaklinks,colorlinks,allcolors=iccvblue]{hyperref}

\def\paperID{7541} 
\def\confName{ICCV}
\def\confYear{2025}

\title{Balancing Task-invariant Interaction and Task-specific Adaptation for Unified Image Fusion}

\author{Xingyu Hu\textsuperscript{1}
\quad
Junjun Jiang\textsuperscript{1}\thanks{Corresponding author.}
\quad
Chenyang Wang\textsuperscript{1}
\quad
Kui Jiang\textsuperscript{1}
\quad
Xianming Liu\textsuperscript{1}
\quad
Jiayi Ma\textsuperscript{2}\\
\textsuperscript{1}Harbin Institute of Technology\quad 
\textsuperscript{2}Wuhan University
\\
{\tt\small \{huxingyu, jiangjunjun, cswcy, jiangkui, csxm\}@hit.edu.cn\quad jyma2010@gmail.com}
}

\begin{document}
\maketitle

\begin{abstract}
Unified image fusion aims to integrate complementary information from multi-source images, enhancing image quality through a unified framework applicable to diverse fusion tasks. While treating all fusion tasks as a unified problem facilitates task-invariant knowledge sharing, it often overlooks task-specific characteristics, thereby limiting the overall performance. Existing general image fusion methods incorporate explicit task identification to enable adaptation to different fusion tasks. However, this dependence during inference restricts the model's generalization to unseen fusion tasks. To address these issues, we propose a novel unified image fusion framework named ``TITA'', which dynamically balances both Task-invariant Interaction and Task-specific Adaptation. For task-invariant interaction, we introduce the Interaction-enhanced Pixel Attention (IPA) module to enhance pixel-wise interactions for better multi-source complementary information extraction. For task-specific adaptation, the Operation-based Adaptive Fusion (OAF) module dynamically adjusts operation weights based on task properties. Additionally, we incorporate the Fast Adaptive Multitask Optimization (FAMO) strategy to mitigate the impact of gradient conflicts across tasks during joint training. Extensive experiments demonstrate that TITA not only achieves competitive performance compared to specialized methods across three image fusion scenarios but also exhibits strong generalization to unseen fusion tasks. The source codes are released at \href{https://github.com/huxingyuabc/TITA}{https://github.com/huxingyuabc/TITA}.

\end{abstract}
  
\section{Introduction}
\label{sec:intro}

A single sensor cannot capture comprehensive information about the imaging scenario. Image fusion addresses this limitation by integrating multi-source data to enhance both the information richness and quality of the image. Over decades of research, image fusion techniques have enabled critical applications in medical diagnosis~\cite{diagnosis}, high-dynamic-range imaging~\cite{hdr,deephdr}, defocus debluring~\cite{bambnet}, and many high-level tasks~\cite{seafusion,unveilingdepth}.

Image fusion encompasses diverse scenarios, 
such as infrared-visible image fusion (IVF), multi-exposure image fusion (MEF), and multi-focus image fusion (MFF). These fusion tasks exhibit distinct characteristics. 
For instance, IVF emphasizes salient thermal features from infrared images while preserving texture details from visible images. MEF balances color and brightness across over- and under-exposed images, whereas MFF focuses on extracting sharp regions from near- and far-focus images.
Despite variations in feature extraction objectives, all image fusion methods share a fundamental goal: integrating complementary information and generating high-quality fused images.

Two types of work have been proposed to cope with diverse scenarios: unified image fusion and general image fusion. 
Unified image fusion methods~\cite{pmgi,u2fusion,fusiondn,uifgan} employ a shared network structure and a universal objective, treating different fusion tasks as a unified problem. While these methods enable task-invariant knowledge transfer, they often neglect task-specific characteristics, ultimately limiting their performance. In contrast, general image fusion methods~\cite{swinfusion,tcmoa} incorporate task-specific characteristics, achieving better adaptability. However, they rely on task identification during inference, 
which restricts the generalization to unseen tasks. Therefore, developing a unified image fusion framework that simultaneously embraces the shared goal of different fusion tasks, \textit{i.e.}, complementary information integration, while incorporating task-specific adaptation remains an unresolved issue.
Furthermore, discrepancies in different fusion tasks can lead to gradient conflicts. This highlights the need for multi-task optimization techniques that ensure balanced performance on diverse fusion tasks. In summary, we identify three key challenges in developing a task-agnostic unified image fusion framework: (i) \textbf{Task-invariant Interaction}: How to effectively explore shared properties (\textit{e.g.}, complementary data integration) across tasks; (ii) \textbf{Task-specific Adaptation}: How to dynamically adapt to task-specific characteristics without explicit task identification; and (iii) \textbf{Multi-objective Optimization}: How to avoid the impact of gradient conflicts in different fusion tasks.

To address these challenges, we try to explore unified image fusion with balanced task-invariant interaction and task-specific adaptation, while aiming to reduce conflicts arising from task differences and enhance overall fusion performance. In this paper, we propose a task-agnostic unified image fusion framework designed to integrate both Task-invariant Interaction and Task-specific Adaptation, named \textbf{TITA}. Specifically, for task-invariant interaction, the Interaction-enhanced Pixel Attention (IPA) block is employed to enhance pixel-level interactions at corresponding spatial locations, improving complementary information extraction, facilitating cross-task cooperation and improving generalization. For task-specific adaptation, we introduce an Operation-Based Adaptive Fusion (OAF) module to dynamically handle task-specific feature fusion requirements. Furthermore, for multi-objective optimization, we 
incorporate the Fast Adaptive Multitask Optimization (FAMO) strategy to dynamically adjust the gradient weights across fusion tasks to mitigate the impact of conflicts between different fusion tasks, ensuring fair optimization.
We summarize the contributions as follows:
\begin{itemize}
    \item We establish a unified image fusion framework that simultaneously explores the task-invariant interactions (through complementary information extraction), and task-specific adaptations, addressing the limitation of prior methods that either ignore the task-specific characteristics or require explicit task identification.
    \item An Interaction-enhanced Pixel Attention (IPA) block that discovers multi-source feature correlations is introduced to explore task-invariant properties. An Operation-based Adaptive Fusion (OAF) module that dynamically adjusts fusion operations is designed to capture task-specific properties without explicit task identification. The framework is reinforced by Fast Adaptive Multitask Optimization (FAMO) strategy to mitigate objective conflicts.
    \item Extensive experiments demonstrate that our framework not only achieves competitive performance to specialized methods across three image fusion scenarios but also exhibits strong generalization to unseen tasks.
\end{itemize}

\section{Related Work}
\label{sec:related}

\subsection{Specialized Image Fusion}

In recent years, specialized image fusion methods have attracted significant attention. Early deep learning-based specialized image fusion methods dedicated to incorporate CNN~\cite{cnn,deepfuse,densefuse,emfusion} and GAN~\cite{fusiongan,mefgan,mffgan} into image fusion pipelines, establishing foundation paradigms. Subsequent breakthroughs introduced physically-informed architectures through deep image priors~\cite{zmff,inpa} to enable zero-shot learning and deep unfolding networks~\cite{ivfnet,dmfusion,derun} to achieve explainable image fusion. 
Researchers have demonstrated the effectiveness of self-supervised image fusion paradigms based on measurement consistency~\cite{emma,refusion}. 
In addition to the advancements in learning paradigms, some studies focused on enhancing perceptual quality, generating fused images with better color and contrast~\cite{piafusion,divfusion}. 
Meanwhile, the combination of image fusion with downstream high-level tasks (such as detection~\cite{tardal,detfusion,metafusion}, segmentation~\cite{segmif,paif,mrfs}, \textit{etc.}) has been well-explored.
With the continuous development of the research frontiers, recent methods have begun to exploit the strong generative capabilities of diffusion models~\cite{ddpm} to produce higher-quality fused images containing richer information~\cite{diffusion,ddfm,diffif,fusiondiff,drmf}.
Furthermore, the rise of large language models has created new opportunities for text-guided image fusion~\cite{textif,textfusion,film}.

\subsection{Unified Image Fusion}

Considering the widespread demand for image fusion and the shared goal of the different fusion tasks (extracting complementary information from multi-source images), many researchers have explored unified image fusion frameworks that can accommodate multiple fusion tasks, leading to significant advancements. 
To unify the objective functions of different fusion tasks, some methods categorize image fusion tasks into two main aspects: structural information preservation and texture detail retention, then adjust the loss function weights to deal with different fusion tasks~\cite{pmgi,u2fusion,fusiondn,uifgan}. 
Subsequent research has further refined these strategies by designing more powerful network architectures and loss functions~\cite{sdnet,lpuif}. Additionally, autoencoder-based image fusion methods have also been applied to unified image fusion frameworks for their ability to learn high-dimensional feature representations while suppressing interference across different modalities~\cite{defusion}. By the way, IFCNN was trained using multi-focus image pairs in a supervised manner and then directly applied to various fusion tasks, demonstrating the generalization due to the shared goals across different fusion scenarios.
Among these advancements, CCF~\cite{ccf} first proposed a sampling-adaptive condition selection mechanism that tailors condition selection at different denoising steps. However, these methods lack consideration of task-specific characteristics and precise objective guidance tailored for each fusion task.

\begin{figure*}[t]
  \centering
    \begin{subfigure}[b]{0.7\linewidth}
        \centering
        \includegraphics[width=\linewidth]{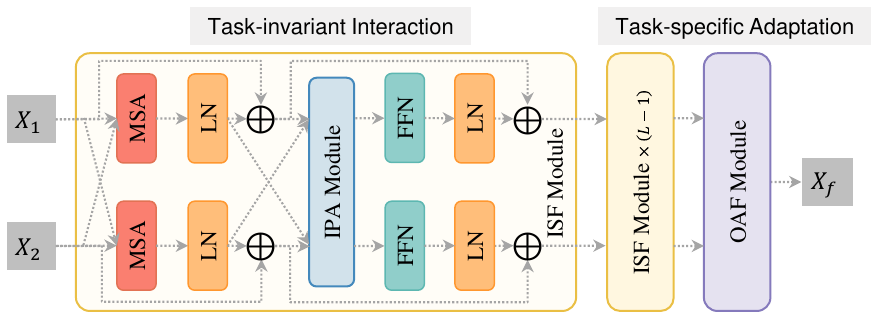}
        \caption{TITA Framework}
        \label{fig:tita}
    \end{subfigure}
    \vspace{0.3cm}
    \begin{subfigure}[b]{0.4\linewidth}
        \centering
        \includegraphics[width=\textwidth]{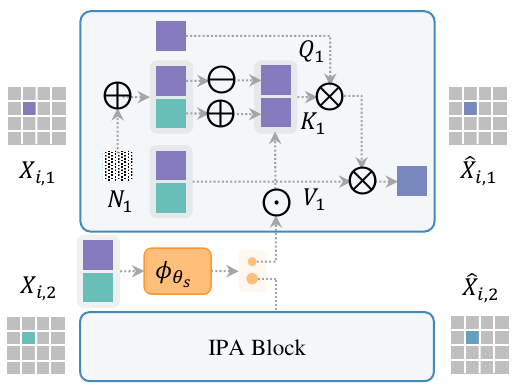}
        \subcaption{IPA Module}
        \label{fig:ipa}
    \end{subfigure}
    \hspace{0.06\linewidth}
    \begin{subfigure}[b]{0.4\linewidth}
        \centering
        \includegraphics[width=\textwidth]{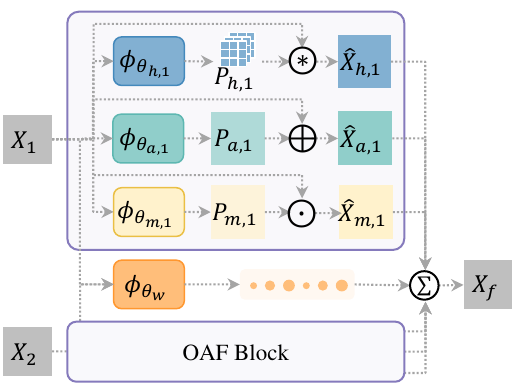}
        \subcaption{OAF Module}
        \label{fig:oaf}
    \end{subfigure}
  \vspace{-5pt}
  \caption{The overall architecture. (a) TITA framework is built on SwinFusion~\cite{swinfusion}, and includes two stages: task-invariant interaction and task-specific adaptation. The task-invariant interaction stage comprises $L$ stacked Interaction-enhanced SwinFusion (ISF) modules, each incorporating an Interaction-enhanced Pixel Attention (IPA) module. The task-specific adaptation is achieved through the proposed Operation-based Adaptive Fusion (OAF) module. (b) The illustration of the IPA module. (c) The illustration of the OAF module.}
  \label{fig:overview}
\end{figure*}

\subsection{General Image Fusion}

In addition, also for the shared goal of the different fusion tasks, some recent general image fusion methods have attempted to take task-specific characteristics into consideration and bridge the gap between specified and unified image fusion methods. SwinFusion~\cite{swinfusion} introduces a general fusion network framework, combining a shared network structure with task-specific training objectives. TC-MoA~\cite{tcmoa}, inspired by the mixture of experts mechanism, proposes a task-specific mixture-of-adapters system to better adapt to different fusion scenarios. While these general image fusion methods have demonstrated outstanding performance across multiple tasks, they still require task identification, which limits their generalization ability to unseen fusion scenarios.

\section{Method}
\label{sec:method}

In this work, we demonstrate that a unified image fusion approach, which effectively integrates both task-invariant interaction and task-specific adaptation, can be achieved without relying on task identification, ensuring adaptability and fusion performance. Given multi-source images $I_1\in\mathbb R^{H\times W\times C_1},I_2\in\mathbb R^{H\times W\times C_2}$ as input, the goal of image fusion is to generate a fused image $I_f\in\mathbb R^{H\times W\times C_f}$, where $H, W, C_1, C_2, C_f$ denote the height, width and the number of channels of source images and the fused image respectively. After tokenization, the input feature are transformed into token sequences $X_{1}, X_{2}\in\mathbb R^{N\times D}$, where $N$ is the number of tokens, $D$ is the embedding dimension. As shown in \cref{fig:overview}, the proposed \textbf{TITA} network including task-invariant interaction and task-specific adaptation two stages, which are introduced in Section \ref{subsec:ti} and Section \ref{subsec:ta} respectively. And the multi-objective optimization strategy is detailed in Section \ref{subsec:mo}.

\subsection{Task-Invariant Feature Interaction}
\label{subsec:ti}
In image fusion tasks, multi-source images contain rich complementary information. The main challenge in capturing this task-invariant characteristic is to effectively extract and utilizing this information, which necessitates enhanced interactions between multi-source features.

\vspace{-10pt}
\paragraph{Revisiting Pixel Attention}

Recent advances in Pixel Attention~\cite{geminifusion} 
show that the interaction between cross-image aligned tokens can be dynamically adjusted. 
Formally, given the $i$-th token $X_{i,1}, X_{i,2}$ of the source feature, the PA mechanism can be formulated as:
\begin{equation}
\begin{aligned}
    &\hat X_{i,1}=MSA(Q_1,K_1,V_1)+X_{i,1},\\
    &\hat X_{i,2}=MSA(Q_2,K_2,V_2)+X_{i,2},\\
    &MSA(Q,K,V)=Softmax(QK^T/\sqrt{D})V,
\end{aligned}
\label{eq:pa1}
\end{equation}
where $Q_1,K_1,V_1$ and $Q_2,K_2,V_2$ can be calculated as:
\begin{equation}
\begin{aligned}
    Q_1&=X_{i,1}W_Q,\\
    K_1&=[(N_K+X_{i,1})W_K,X_{i,1}\phi_{\theta_s}(X_{i,1},X_{i,2})W_K],\\
    V_1&=[(N_V+X_{i,1})W_V,X_{i,2}W_V],\\
    Q_2&=X_{i,2}W_Q,\\
    K_2&=[(N_K+X_{i,2})W_K,X_{i,2}\phi_{\theta_s}(X_{i,2},X_{i,1})W_K],\\
    V_2&=[(N_V+X_{i,2})W_V,X_{i,1}W_V],
\end{aligned}
\label{eq:pa2}
\end{equation}
where $\phi_{\theta_s}(\cdot)$ represents the relation discriminator (a 2-layer MLP with a sigmoid activation). The PA mechanism essentially performs adaptive weight allocation between self-attention and cross-attention operations. The layer-adaptive noise $N_K,N_V$ are introduced to counteract the self-attention bias, as self-attention operations inherently tend to produce higher attention scores than cross-attention operations.

\vspace{-5pt}
\paragraph{Interaction-enhanced Pixel Attention}

Motivated by PA, we proposed an Interaction-enhanced Pixel Attention (IPA) module to enhance the interaction of multi-source images to extract complementary information. Two major modifications has been made based on PA module. First, we remove the direct noise injection to $V$, as it may cause irreversible information loss and accuracy degradation. Second, to encourage a stronger preference for cross-attention operations to emphasize complementary information integration. The illustration is shown in \cref{fig:ipa}. IPA is can be calculated as:
\begin{equation}
\begin{aligned}
    K_1=&[(N_1+X_{i,1}-X_{i,1}\phi_{\theta_s}(X_{i,1},X_{i,2}))W_K,\\&(N_2+X_{i,1}+X_{i,1}\phi_{\theta_s}(X_{i,1},X_{i,2}))W_K],\\
    V_1=&[X_{i,1}W_V,X_{i,2}W_V],\\
    K_2=&[(N_1+X_{i,2}-X_{i,2}\phi_{\theta_s}(X_{i,2},X_{i,1}))W_K,\\&(N_2+X_{i,2}+X_{i,2}\phi_{\theta_s}(X_{i,2},X_{i,1}))W_K],\\
    V_2=&[X_{i,2}W_V,X_{i,1}W_V].
\end{aligned}
\label{eq:ipa}
\end{equation}
This design thus establishes an explicit causal relationship: higher irrelevance scores estimated by the relation discriminator $\phi_{\theta_s}(\cdot)$ directly amplify cross-attention weights, as detailed in \cref{subsubsec:ti}. Moreover, this asymmetric design intentionally reinforces cross-attention, ensuring that complementary interactions are fully leveraged.

Moreover, we modify the SwinFusion module by replacing all intra-domain fusion with inter-domain fusion, introducing the Interaction-enhanced SwinFusion (ISF) module, as shown in \cref{fig:overview}. This design further strengthens multi-source image interactions by increasing the number of cross-attention operations.

\subsection{Task-Specific Feature Fusion}
\label{subsec:ta}

Different image fusion tasks exhibit distinct physical characteristics, imposing different requirements on feature fusion. To accommodate these task-specific properties, we design an Operation-based Adaptive Fusion (OAF) module that dynamically modulates the weights of operation branches based on each task's physical properties.

Specifically, the OAF module consists of three parallel operation branches: (i) \textbf{HPF} branch: applies spatially-variant high-pass filtering to capture high-frequency details, enhancing texture and edge information. (ii) \textbf{ADD} branch: performs residual addition for overall information enhancement. (iii) \textbf{MUL} branch: performs element-wise multiplication to facilitate nonlinear feature interactions to capture complex feature relationships. (For details, please refer Appendix A.1.) The dynamic weights of these branches reflect the physical characteristics of different fusion tasks. Instead of manually assigning fixed weights to each fusion task, we employ a dynamic weight prediction network that takes task-specific features as input and predicts adaptive weights to determine the optimal weight distribution. As illustrated in \cref{fig:oaf}, given multi-source features $X_1,X_2$, the hypernetworks generate operands $P_h=\phi_{\theta_h}(X), P_a=\phi_{\theta_a}(X), P_m=\phi_{\theta_m}(X)$. And the resultant features $\hat X_h=X*P_h, \hat X_a=X+P_a,\hat X_m=X\odot P_m$ are then dynamically weighted by the weights $W=\phi_{\theta_w}(X_1,X_2)$.
Then the output feature $X_f$ can be obtained as:
\begin{equation}
\begin{aligned}
    X_f=\sum_{o\in\{h,a,m\}}{W_1\cdot \hat X_{o,1}+W_2\cdot \hat X_{o,2}}.
\end{aligned}
\end{equation}
In this way, explicit task identification is eliminated, enabling task-agnostic feature fusion.

\begin{figure}[t]
  \centering
   \includegraphics[width=1\linewidth]{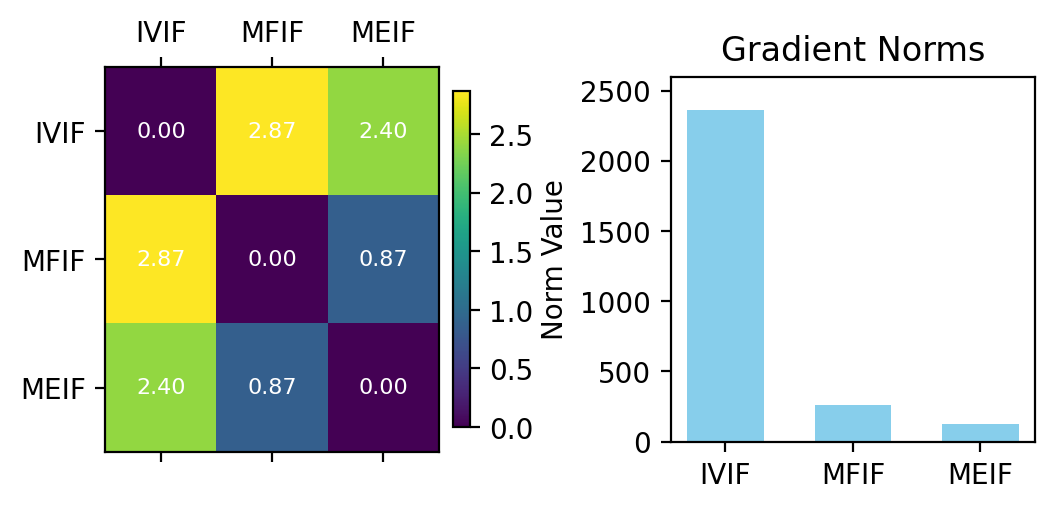}
   \caption{The gradient conflict angles and the gradient norms on three fusion tasks when TITA operates without multi-objective optimization. We observed gradient conflicts occurring frequently during training, and the case shown here (iteration=$3000$) was selected as a representative example.}
   \label{fig:gradient}
\end{figure}

\begin{figure*}[]
  \centering
   \includegraphics[width=1\linewidth]{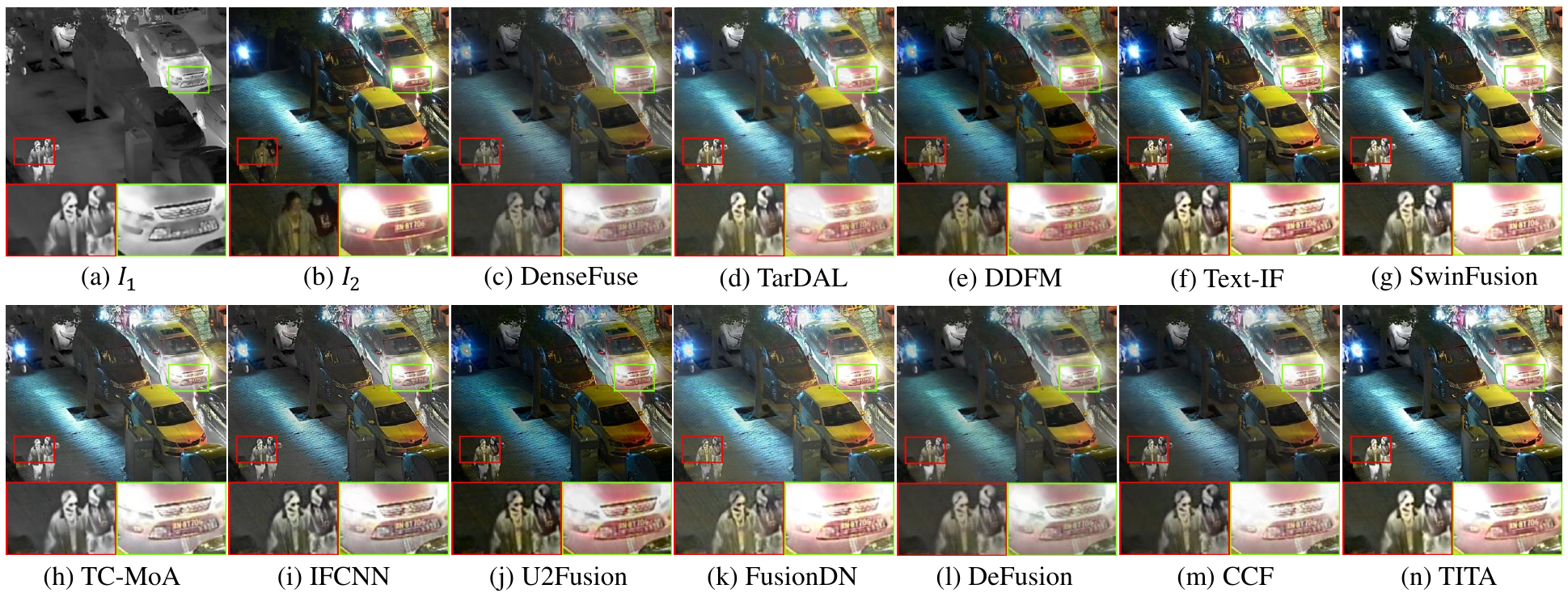}
   \caption{Visual comparisons of SOTA approaches for IVF task.}
   \label{fig:quanlitative_ivf}
\end{figure*}

\subsection{Multi-objective Optimization}
\label{subsec:mo}

To enhance performance of each single task within our framework, we follow previous general image fusion methods~\cite{swinfusion} to use task-specific objectives, the detailed introduction is given in Appendix A.2. We use task identity only during training, while for inference, the proposed task-specific adaptation accommodates different fusion tasks, ensuring compliance with the task-agnostic constraint.

Formally, considering $M$ fusion tasks associated with $M$ objectives $\{\ell_m(\theta)\}_{m=1}^M$, where $\theta$ is the parameter of unified model. The direct way to optimize $\theta$ is to define a single objective by averaging objectives of tasks: $\ell(\theta)=1/M\sum_{m=1}^M  L_m(\theta)$. However, as shown in \cref{fig:gradient}, we observe conflicting gradients and large difference in gradient magnitudes in the training, which would cause detrimental impact on the optimization ~\cite{liu2021conflict}. Thus, we instead view the problem as a multi-objective optimizing problem by defining $L(\theta)=(\ell_1(\theta),\ell_2(\theta),\cdots,\ell_M(\theta))$ and turning to seek the Pareto optimal point \cite{miettinen1999nonlinear} $\theta^*$ of $L(\theta)$. We adopt FAMO~\cite{famo} as our multi-objective optimizing algorithm for the following reasons: (i) it is easy to implement; (ii) it introduces negligible computation overhead; (iii) the loss reduce rates of different tasks equal in FAMO (with some constraints). Please refer to the Appendix A.3 for more details about FAMO.

\section{Experiments}
\label{sec:experiment}

\subsection{Experimental Setup}

\paragraph{Implement Details}

We use an Adam optimizer with learning rate set to $2e^{-5}$ for model parameter optimization, and an Adam optimizer with learning rate set to $0.025$ and weight decay set to $1e^{-3}$ for weighting logits optimization (introduced by FAMO). The batch size is set to $8$ and the number of iterations is set to $20000$. Training data is unevenly distributed across fusion tasks due to varying acquisition costs and difficulties. To address this imbalance, we uniformly sample data from each task for each iteration.
The proposed method is implemented by Pytorch and all experiments are conducted on a device equipped with a NVIDIA RTX 3090 GPU. 

\paragraph{Datasets}

We follow TC-MoA~\cite{tcmoa} to construct the training and testing datasets. The training dataset consists of $12025$ infrared and visible image pairs from \textit{LLVIP} dataset~\cite{llvip}, $589$ multi-exposure image pairs from \textit{SCIE} dataset~\cite{sice}, $710$ multi-focus image pairs from \textit{RealMFF} dataset~\cite{realmff}, and $90$ multi-focus image pairs from \textit{MFI-WHU} dataset~\cite{mfiwhu}. Although the trainig data size varies across different fusion tasks, all tasks are sampled equally to mitigate the imbalance. For evaluation on IVF, the testing dataset includes $70$ infrared and visible image pairs from \textit{LLVIP} dataset~\cite{llvip}. For evaluation on MEF, the testing dataset includes $100$ multi-exposure image pairs from \textit{MEFB} dataset~\cite{mefb}. And for evaluation on MFF, the testing dataset contains $20$ multi-focus image pairs from \textit{Lytro} dataset~\cite{lytro}, $13$ multi-focus image pairs from \textit{MFFW} dataset~\cite{mffw}, and $30$ multi-focus image pairs from \textit{MFI-WHU} dataset~\cite{mfiwhu}. Besides, we validate the generalization of the proposed unified image fusion method on \textit{Harvard} dataset \footnote{\href{http://www.med.harvard.edu/AANLIB/home.html}{http://www.med.harvard.edu/AANLIB/home.html}} for medical image fusion, and \textit{Quickbird} dataset \footnote{\href{https://www.satimagingcorp.com/satellite-sensors/quickbird/}{https://www.satimagingcorp.com/satellite-sensors/quickbird/}} for pan-sharpening following PMGI~\cite{pmgi}.

\vspace{-5pt}

\paragraph{Metrics}

Many non-reference metrics have been proposed to better evaluate fusion results. According to~\cite{survey}, these metrics can be categorized into four groups: information theory-based, image feature-based, structural similarity-based, and human perception-based. Information theory based metrics, such as MI and FMI, measure the amount of mutual information preserved from source images; Image feature based metrics, including SD, Qabf and EI, assess the fused image quality based on image features; Structural similarity based metrics, including Qw and MEF-SSIM, evaluate the structural similarity between the fused image and source images; and human perception based metric, such as VIF, estimate the fusion quality from a human perceptual perspective.

\vspace{-5pt}

\paragraph{Methods for comparison}

For better evaluation, we compare our methods with both unified image fusion methods including IFCNN~\cite{ifcnn}, PMGI~\cite{pmgi}, U2Fusion~\cite{u2fusion}, FusionDN~\cite{fusiondn}, DeFusion~\cite{defusion}, CCF~\cite{ccf} and general image fusion methods including SwinFusion~\cite{swinfusion}, TC-MoA~\cite{tcmoa}. Besides, we also compare with some representative task-specific image fusion methods. For IVF, we select DenseFuse~\cite{densefuse}, TarDAL~\cite{tardal}, DDFM~\cite{ddfm}, Text-IF~\cite{textfusion}. For MEF, we compare with MEF-GAN~\cite{mefgan}, and HoLoCo~\cite{holoco}. For MFF, we compare with MFF-GAN~\cite{mffgan} and note that IFCNN is originally trained in a supervised manner for the MFF task.

\begin{figure}[]
  \centering
   \includegraphics[width=1\linewidth]{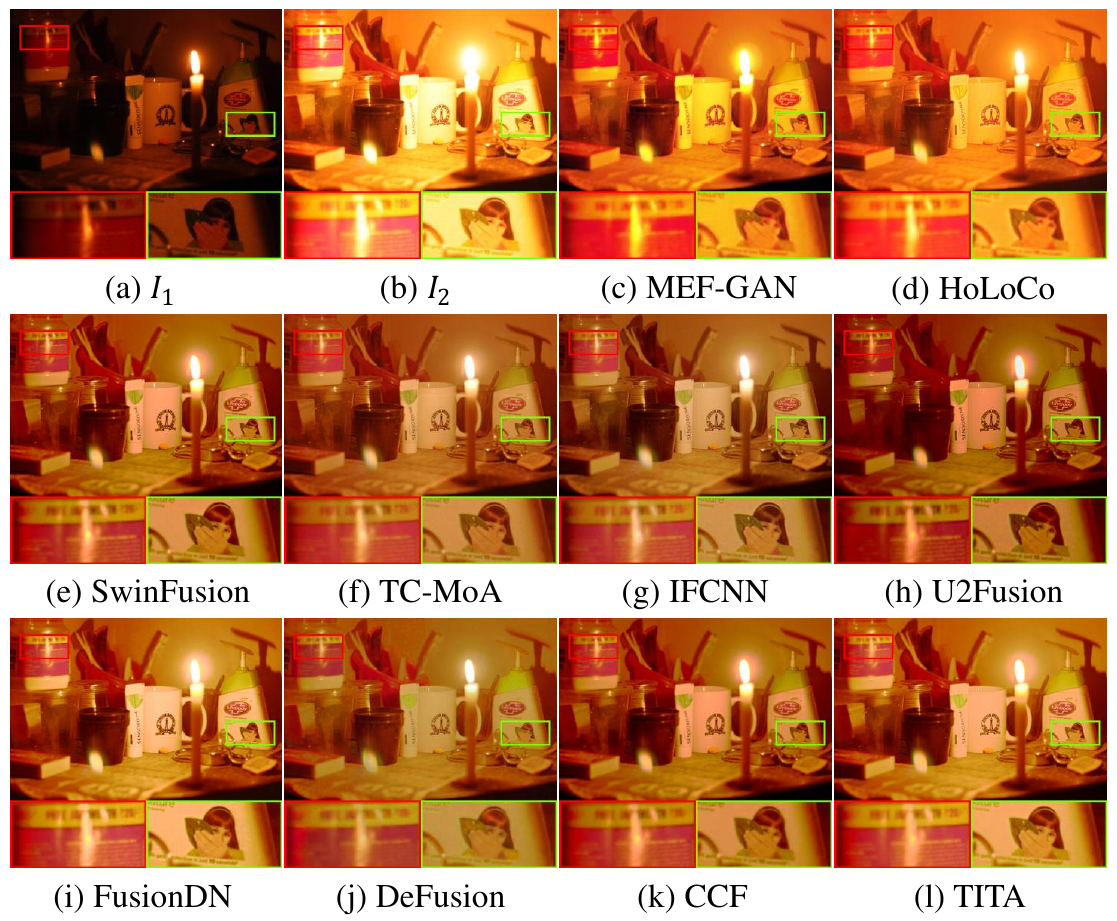}
   \caption{Visual comparisons of SOTA approaches for MEF task.}
   \label{fig:quanlitative_mef}
\end{figure}

\begin{figure}[]
  \centering
   \includegraphics[width=1\linewidth]{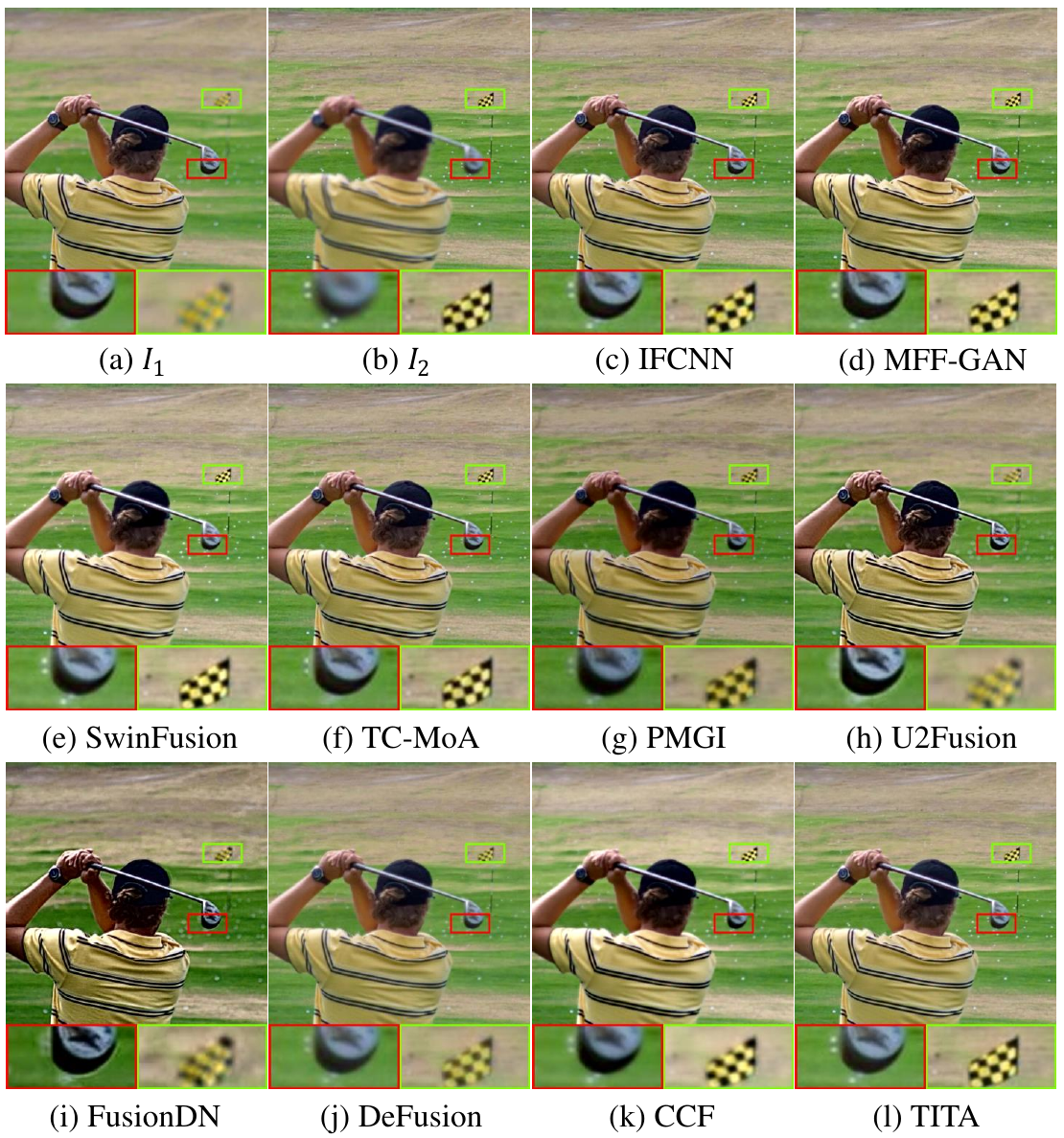}
   \caption{Visual comparisons of SOTA approaches for MFF task.}
   \label{fig:quanlitative_mff}
\end{figure}

\begin{figure*}[]
  \centering
   \includegraphics[width=1\linewidth]{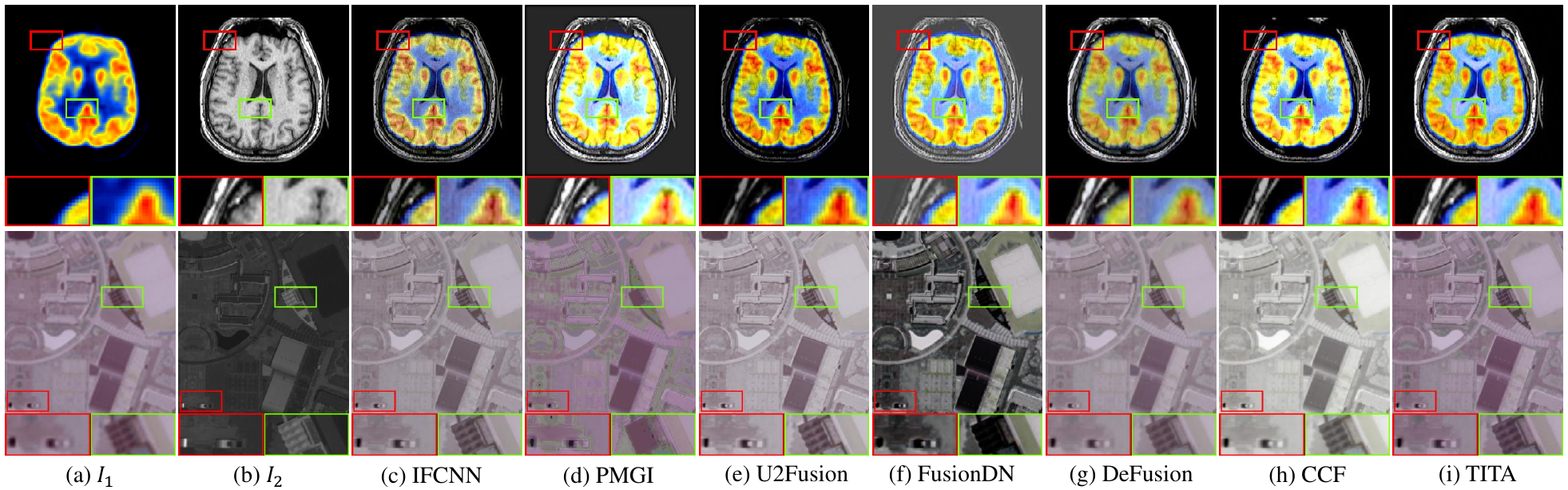}
   \caption{The generalization results for MIF and PAN task.}
   \label{fig:generalization}
\end{figure*}

\begin{table}[t]
\centering
\caption{The quantitative results for IVF task. (\textbf{Bold}: best; \ul{Underline}: second best.)}
\footnotesize
\begin{tabular}{@{}lllllll@{}}
\toprule
Method                                            & MI             & FMI            & Qabf           & Qp             & Qw             & VIF            \\ \midrule
\multicolumn{7}{c}{Specialized}                                                                                                                        \\ \midrule
\multicolumn{1}{l|}{DenseFuse~\cite{densefuse}}   & 2.687          & 0.877          & 0.317          & 0.309          & 0.559          & 0.675          \\
\multicolumn{1}{l|}{TarDAL~\cite{tardal}}         & 3.152          & 0.862          & 0.413          & 0.285          & 0.614          & 0.710          \\
\multicolumn{1}{l|}{DDFM~\cite{ddfm}}             & 2.921          & 0.884          & 0.507          & 0.427          & 0.686          & 0.749          \\
\multicolumn{1}{l|}{Text-IF~\cite{textif}}        & 3.322          & {\ul 0.892}    & \textbf{0.684} & {\ul 0.465}    & {\ul 0.859}    & \textbf{0.932} \\ \midrule
\multicolumn{7}{c}{General}                                                                                                                             \\ \midrule
\multicolumn{1}{l|}{SwinFusion~\cite{swinfusion}} & 3.873          & 0.889          & 0.650          & 0.457          & 0.844          & 0.907          \\
\multicolumn{1}{l|}{TC-MoA~\cite{tcmoa}}          & {\ul 3.606}    & 0.886          & 0.600          & 0.392          & 0.855          & 0.925          \\ \midrule
\multicolumn{7}{c}{Unified}                                                                                                                             \\ \midrule
\multicolumn{1}{l|}{IFCNN~\cite{ifcnn}}           & 2.821          & 0.878          & 0.582          & 0.361          & 0.840          & 0.773          \\
\multicolumn{1}{l|}{PMGI~\cite{pmgi}}             & 3.096          & 0.869          & 0.176          & 0.223          & 0.333          & 0.544          \\
\multicolumn{1}{l|}{U2Fusion~\cite{u2fusion}}     & 2.447          & 0.870          & 0.378          & 0.303          & 0.625          & 0.576          \\
\multicolumn{1}{l|}{FusionDN~\cite{fusiondn}}     & 2.666          & 0.871          & 0.494          & 0.328          & 0.753          & 0.758          \\
\multicolumn{1}{l|}{DeFusion~\cite{defusion}}     & 2.687          & 0.877          & 0.317          & 0.316          & 0.676          & 0.675          \\
\multicolumn{1}{l|}{CCF~\cite{ccf}}               & 2.789          & 0.881          & 0.499          & 0.347          & 0.726          & 0.719          \\
\multicolumn{1}{l|}{TITA (Ours)}                         & \textbf{4.176} & \textbf{0.896} & {\ul 0.679}    & \textbf{0.473} & \textbf{0.877} & {\ul 0.926}    \\ \bottomrule
\end{tabular}
\label{tab:com_ivf}
\end{table}

\begin{table}[]
\centering
\caption{The quantitative results for MEF task. (\textbf{Bold}: best; \ul{Underline}: second best.)}
\footnotesize
\begin{tabular}{@{}lllllll@{}}
\toprule
Method                                            & MI             & FMI            & SD             & EI             & Qw             & VIF            \\ \midrule
\multicolumn{7}{c}{Specialized}                                                                                                                        \\ \midrule
\multicolumn{1}{l|}{MEF-GAN~\cite{mefgan}}        & 4.844          & 0.874          & 10.26          & 45.96          & 0.589          & 1.371          \\
\multicolumn{1}{l|}{HoLoCo~\cite{holoco}}         & 4.644          & 0.874          & 10.37          & 49.85          & 0.690          & 1.463          \\ \midrule
\multicolumn{7}{c}{General}                                                                                                                             \\ \midrule
\multicolumn{1}{l|}{SwinFusion~\cite{swinfusion}} & 5.287          & \textbf{0.901} & 10.26          & {\ul 54.45}    & \textbf{0.918} & 1.463          \\
\multicolumn{1}{l|}{TC-MoA~\cite{tcmoa}}          & 5.505          & 0.890          & 10.23          & 49.12          & 0.841          & 1.485          \\ \midrule
\multicolumn{7}{c}{Unified}                                                                                                                             \\ \midrule
\multicolumn{1}{l|}{IFCNN~\cite{ifcnn}}           & 5.695          & 0.888          & 10.37          & 53.94          & 0.848          & 1.436          \\
\multicolumn{1}{l|}{PMGI~\cite{pmgi}}             & 5.093          & 0.878          & 10.28          & 35.12          & 0.597          & 1.086          \\
\multicolumn{1}{l|}{U2Fusion~\cite{u2fusion}}     & 5.526          & 0.882          & 10.14          & 43.19          & 0.778          & 1.301          \\
\multicolumn{1}{l|}{FusionDN~\cite{fusiondn}}     & 5.762          & 0.880          & 10.46          & 49.90          & 0.763          & 1.403          \\
\multicolumn{1}{l|}{DeFusion~\cite{defusion}}     & 4.626          & 0.882          & {\ul 10.46}    & 42.05          & 0.758          & 1.159          \\
\multicolumn{1}{l|}{CCF~\cite{ccf}}               & {\ul 5.879}    & 0.885          & 10.44          & 39.48          & 0.703          & {\ul 1.515}    \\
\multicolumn{1}{l|}{TITA (Ours)}                         & \textbf{6.207} & {\ul 0.900}    & \textbf{10.73} & \textbf{55.07} & {\ul 0.891}    & \textbf{1.534} \\ \bottomrule
\end{tabular}
\label{tab:com_mef}
\end{table}

\begin{table}[]
\centering
\caption{The quantitative results for MFF task. (\textbf{Bold}: best; \ul{Underline}: second best.)}
\footnotesize
\begin{tabular}{@{}lllllll@{}}
\toprule
Method                                            & MI             & FMI            & Qabf           & SSIM       & Qw             & VIF            \\ \midrule
\multicolumn{7}{c}{Specialized}                                                                                                                        \\ \midrule
\multicolumn{1}{l|}{IFCNN~\cite{ifcnn}}           & 6.495          & {\ul 0.882}    & 0.658          & 0.991          & \textbf{0.912} & 1.614          \\
\multicolumn{1}{l|}{MFF-GAN~\cite{mffgan}}        & 5.749          & 0.875          & 0.628          & 0.980          & 0.892          & 1.468          \\ \midrule
\multicolumn{7}{c}{General}                                                                                                                             \\ \midrule
\multicolumn{1}{l|}{SwinFusion~\cite{swinfusion}} & 6.261          & 0.881          & {\ul 0.687}    & {\ul 0.991}    & 0.893          & 1.633          \\
\multicolumn{1}{l|}{TC-MoA~\cite{tcmoa}}          & \textbf{6.695} & 0.881          & 0.600          & 0.990          & 0.891          & \textbf{1.655} \\ \midrule
\multicolumn{7}{c}{Unified}                                                                                                                             \\ \midrule
\multicolumn{1}{l|}{PMGI~\cite{pmgi}}             & 5.511          & 0.866          & 0.351          & 0.902          & 0.523          & 1.340          \\
\multicolumn{1}{l|}{U2Fusion~\cite{u2fusion}}     & 5.219          & 0.869          & 0.469          & 0.902          & 0.518          & 1.392          \\
\multicolumn{1}{l|}{FusionDN~\cite{fusiondn}}     & 5.451          & 0.863          & 0.434          & 0.880          & 0.485          & 1.407          \\
\multicolumn{1}{l|}{DeFusion~\cite{defusion}}     & 5.833          & 0.870          & 0.472          & 0.960          & 0.693          & 1.569          \\
\multicolumn{1}{l|}{CCF~\cite{ccf}}               & 5.823          & 0.875          & 0.474          & 0.952          & 0.651          & 1.604          \\
\multicolumn{1}{l|}{TITA (Ours)}                         & {\ul 6.546}    & \textbf{0.885} & \textbf{0.697} & \textbf{0.993} & {\ul 0.907}    & {\ul 1.637}    \\ \bottomrule
\end{tabular}
\label{tab:com_mff}
\end{table}

\subsection{Comparison with State-of-the-arts}

\subsubsection{Results on Multi-Modal Image Fusion}

The quantitative comparison results for IVF on \textit{LLVIP} dataset are shown in \cref{tab:com_ivf}. Overall, TITA outperforms both general and unified methods, and provides comparable results to specialized methods, highlighting its effectiveness on IVF task. Specifically, our method demonstrates superior results in terms of MI, FMI, Qp, and Qw, indicating that it effectively preserves and integrates more information from the source images. Meanwhile, the specialized fusion method Text-IF achieves the best perceptual performance in Qabf and VIF due to the advantage of a powerful pretrained LLM. From the visual comparison results provided in \cref{fig:quanlitative_ivf}, both TITA and SwinFusion perform well in preserving salient objects and texture details. However, TITA exhibits better contrast than SwinFusion, benefiting from its strong learning capability across multiple fusion tasks.

\subsubsection{Results on Multi-Exposure Image Fusion}

As shown in \cref{tab:com_mef}, TITA achieves the best performance on MI, SD, EI, and VIF, indicating its strong ability to retain source image information and produce visually pleasing results. The general image fusion method SwinFusion shows great results on FMI and Qw due to the task-specific training on MEF task. Notably, TITA outperforms other unified methods, validating its effectiveness for the MEF task. As shown in \cref{fig:quanlitative_mef}, TITA excels in detail preservation and contrast enhancement.

\subsubsection{Results on Multi-Focus Image Fusion}

As shown in \cref{tab:com_mff}, TITA achieves the highest scores on FMI, Qabf, and MEF-SSIM, demonstrating its strong ability to preserve source information and structural details. TC-MoA performs well on MI and VIF. However, it requires explicit task identification and post-processing, making it less flexible. Additionally, IFCNN achieves good performance on Qw. As a specialized method trained on MFF task, it performs better than those on other fusion tasks. The qualitative comparison results for MFF are shown in \cref{fig:quanlitative_mff}, demonstrating that the proposed TITA possesses a strong information integration capability in terms of clarity. Specifically, IFCNN, SwinFusion, and our TITA effectively preserve fine details and enhance edge sharpness. However, at the focus boundary, MFF-GAN introduces noticeable artifacts, TC-MoA fails to recover intricate details, and other methods struggle to provide clear and well-defined structures, leading to blurred or distorted fusion results.

\subsubsection{Generalization on Other Fusion Tasks}

We test TITA on two unseen fusion tasks, that is, medical image fusion (MIF) and pan-sharpening (PAN), and compare with six unified image fusion methods. Note that for PMGI~\cite{pmgi}, these two tasks are known. \cref{fig:generalization} shows that TITA demonstrates strong adaptability to different types of fusion scenarios, benefiting from its task-agnostic design. In contrast, FusionDN and CCF fail to generalize and collapse when encountering unseen tasks. This further verifies that the common goal of maximizing the amount of information among different fusion tasks enables the unified method to be effectively generalized to various scenarios.

\begin{table}[]
\footnotesize
\caption{The ablation study on three main components.}
\begin{tabular}{@{}lll|lllll@{}}
\toprule
TI         & TA         & MO         & MI              & FMI             & Qabf            & VIF             & Params \\ \midrule
\multicolumn{3}{c|}{Baseline}        & 2.9632          & 0.8841          & 0.4581          & 0.7027          & 0.97M  \\
\multicolumn{3}{c|}{Baseline-TS}     & 3.6116          & 0.8887          & 0.6464          & 0.8448          & 0.97M  \\
\checkmark &            &            & 3.6853          & 0.8893          & 0.6509          & 0.8546          & 1.30M  \\
           & \checkmark &            & 3.8822          & 0.8922          & 0.6639          & 0.9038          & 1.06M  \\
           &            & \checkmark & 3.6802          & 0.8909          & 0.6619          & 0.8531          & 0.97M  \\
\checkmark & \checkmark &            & 3.8832          & 0.8928          & 0.6663          & 0.9061          & 1.39M  \\
\checkmark &            & \checkmark & 3.7546          & 0.8920          & 0.6671          & 0.8650          & 1.30M  \\
           & \checkmark & \checkmark & 4.1220          & 0.8948          & 0.6763          & 0.9191          & 1.06M  \\
\multicolumn{3}{c|}{Ours}            & \textbf{4.1759} & \textbf{0.8959} & \textbf{0.6795} & \textbf{0.9264} & 1.39M  \\ \bottomrule
\end{tabular}
\label{tab:ablation_ivf}
\end{table}

\subsection{Model Analysis}

\subsubsection{Ablation Study}
\label{subsubsec:ablation}

Since the introduced task-invariant integration (TI), task-specific adaptation (TA), and multi-objective optimization (MO) strategies can function independently, we can easily perform ablation experiments to verify their effectiveness. To establish the baseline, we unified-trained SwinFusion~\cite{swinfusion} across all fusion tasks: (i) \textbf{Baseline}: using unified loss function borrowed from U2Fusion~\cite{u2fusion}; (ii) \textbf{Baseline-TS}: using task-specific training objectives with uniformly sampled task data. For detailed experimental setting of the used objectives, please refer to Appendix A.2.

Several key insights can be drawn from the results in \cref{tab:ablation_ivf}: (i) Incorporating task-specific training objectives improves the performance for a large margin since it provides more precise guidance for each task compared to the unified objective. (ii) Each component contributes to improving the baseline performance, validating their effectiveness. (iii) MO enhances performance on IVF task (as well as the MEF and MFF tasks as shown in the Appendix B.1), while introducing only negligible additional parameters. This improvement suggests the presence of gradient conflicts among different fusion tasks, and the effectiveness of FAMO in alleviating these conflicts. (iv) TA significantly benefits from the inclusion of MO, suggesting that TA introduces gradient conflicts, and MO effectively alleviates this issue. (v) The combination of TI, TA, and MO achieves the best performance, suggesting that these components mutually reinforce each other and enhance the model’s ability to integrate complementary information, resolve gradient conflicts, and adapt to diverse fusion tasks more effectively.

\begin{table}[]
\centering
\footnotesize
\caption{The ablation study on task-invariant integration.}
\begin{tabular}{@{}l|lllll@{}}
\toprule
     & MI     & FMI    & Qabf   & VIF    & Params \\ \midrule
SF   & 3.6116 & 0.8887 & 0.6464 & 0.8448 & 0.97M  \\
IrSF & 3.5334 & 0.8880 & 0.6400 & 0.8330 & 0.97M  \\
IeSF & \textbf{3.6331} & \textbf{0.8889} & \textbf{0.6494} & \textbf{0.8454} & 0.97M  \\ \midrule
TE   & 3.6497 & 0.8885 & 0.6458 & 0.8470 & 1.02M  \\
PA   & 3.6694 & 0.8892 & 0.6485 & 0.8521 & 1.14M  \\
IPA  & \textbf{3.6778} & \textbf{0.8895} & \textbf{0.6515} & \textbf{0.8558} & 1.14M  \\ \bottomrule
\end{tabular}
\label{tab:ti}
\end{table}
\subsubsection{Analysis on Task-invariant Integration}
\label{subsubsec:ti}

This paragraph mainly focuses on the verification of the proposed IPA module and the ISF module. First, we compare the proposed Interaction-enhanced (IeSF) module which replaces all the intra-domain fusion with the inter-domain fusion, with the baseline-ST (with original SwinFusion (SF) module) and the Interaction-reduced (IrSF) module which replaces all the inter-domain fusion with the intra-domain fusion. The key difference among these modules lies in the balance between cross-attention and self-attention operations. Specifically, the number of cross-attention mechanisms follows the order: IeSF $>$ SF $>$ IrSF, highlighting the increasing level of multi-source interactions. The results shown in \cref{tab:ti} indicate that increasing the amount of cross-attention operations leads to improved performance. (Results on MEF and MFF tasks in Appendix B.2 also validate this.) Then we compare IPA with PA~\cite{geminifusion} and Token Exchange (TE)~\cite{tokenfusion}, which is introduced in the Appendix A.4. The results shown in \cref{tab:ti} demonstrate the effectiveness of the proposed IPA module.

\begin{table}[]
\centering
\footnotesize
\caption{The ablation study on task-specific adaptation.}
\begin{tabular}{@{}l|lllll@{}}
\toprule
        & MI     & FMI    & Qabf   & VIF    & Params \\ \midrule
W/o HPF & 3.9688 & 0.8933 & 0.6700 & 0.9114 & 1.38M  \\
W/o ADD & 3.8039 & 0.8952 & 0.6699 & 0.8917 & 1.32M  \\
W/o MUL & 3.8994 & 0.8933 & 0.6641 & 0.8971 & 1.32M  \\
W/o DW  & 3.7844 & 0.8951 & 0.6658 & 0.8711 & 1.38M  \\
Ours    & \textbf{4.1759} & \textbf{0.8959} & \textbf{0.6795} & \textbf{0.9264} & 1.39M  \\ \bottomrule
\end{tabular}
\label{tab:ta}
\end{table}

\subsubsection{Analysis on Task-specific Adaptation}
\label{subsubsec:ta}

To illustrate the importance of each operation branch and the dynamic weight prediction (DW) in the OAF module, we conduct an ablation study, as shown in \cref{tab:ta}. The results indicate that both the DW strategy and the three operation branches significantly contribute to the overall performance, with the MUL branch being the most crucial, which is reasonable given that image fusion inherently involves many non-linear operations. This highlights the necessity of incorporating diverse operations to enhance adaptive feature fusion. Results for MEF and MFF tasks, and visualizations of DW are provided in Appendix B.3 and B.4.

\vspace{-5pt}
\section{Conclusion}
\label{sec:conclusion}

In this work, we propose a unified image fusion framework that effectively integrates both task-invariant and task-specific properties to enhance fusion performance and generalization across different fusion tasks. 
Future work will explore further improvements through dynamic expansion to other unseen fusion scenarios and more modalities, and integration of advanced learning techniques to develop stronger cross-modal representations, enhancing fusion quality and adaptability across diverse applications.

\section*{Acknowledgements}
\label{sec:ackn}

The research was supported by the National Natural Science Foundation of China (U23B2009, 62471158).

{
    \small
    \bibliographystyle{ieeenat_fullname}
    \bibliography{main}
}

\clearpage
\appendix

\section{Supplementary Details}

\subsection{Additional Details of Task-specific Adaptation}

For the details of three parallel operation branches in OAF module: (i) \textbf{HPF} branch: Conditioned by the operand $P_h\in \mathbb R^{\frac{N}{H\times W}\times K^2\times H\times W}$ predicted by the hyper-network $\phi_{\theta_h}(\cdot)$, this branch employs spatially-variant convolution on the input feature $X\in\mathbb R^{\frac{N}{H\times W}\times D\times H\times W}$ to obtain the resultant feature $\hat X_h\in\mathbb R^{\frac{N}{H\times W}\times D\times H\times W}$, where $H, W, K$ denote the height, width and the kernel size respectively. (ii) \textbf{ADD} branch: Conditioned by the operand $P_a\in\mathbb R^{\frac{N}{H\times W}\times D\times H\times W}$ predicted by the hyper-network $\phi_{\theta_a}(\cdot)$, this branch performs addition on the input feature to obtain the resultant feature $\hat X_a\in\mathbb R^{\frac{N}{H\times W}\times D\times H\times W}$. (iii) \textbf{MUL} branch: Conditioned by the operand $P_m\in\mathbb R^{\frac{N}{H\times W}\times D\times H\times W}$ predicted by the hyper-network $\phi_{\theta_m}(\cdot)$, this branch performs Hadamard (element-wise) multiplication on the input feature to obtain the resultant feature $\hat X_m\in\mathbb R^{\frac{N}{H\times W}\times D\times H\times W}$. And all the hyper-networks used are two-layer MLPs.

\subsection{Optimization Objectives}

The unified objective used in Baseline is borrowed from U2Fusion~\cite{u2fusion}:
\begin{equation}
\begin{aligned}
    \ell=&\lambda_1\ell_{ssim}+\lambda_2\ell_{mse},\\
\ell_{ssim}=&w_1(1-ssim(I_f,I_1))+w_2(1-ssim(I_f,I_2)),\\
\ell_{mse}=&w_1\cdot\Vert I_f-I_1 \Vert_2^2+w_2\Vert I_f-I_2\Vert_2^2,
\end{aligned}
\end{equation}
where $\lambda_1=1, \lambda_2=20$, and $w_1,w_2$ is calculated by the information measured on VGG features. 

And for Baseline-TS and the proposed TITA framework, following SwinFusion~\cite{swinfusion}, the task-specific training objectives are shown as below:
\begin{equation}
\begin{aligned}
    \ell=&\lambda_1\ell_{ssim}+\lambda_2\ell_{text}+\lambda_3\ell_{int},\\
\ell_{ssim}=&\frac{1}{2}(1-ssim(I_f,I_1))+\frac{1}{2}(1-ssim(I_f,I_2)),\\
\ell_{text}=&\frac{1}{HW}\Vert \vert\triangledown I_f \vert -\max{(\vert \triangledown I_1 \vert , \vert \triangledown I_2 \vert)}\Vert_1,\\
\ell_{int}=&\frac{1}{HW}\Vert I_f-M(I_1,I_2) \Vert_1,
\end{aligned}
\label{eq:loss}
\end{equation}
where $\lambda_1=10,\lambda_2=20,\lambda_3=20$ are hyper-parameters, $M(\cdot)$ is task-specific element-wise aggregation operation. Specifically, $max(\cdot,\cdot)$ is employed for IVF and MFF, $mean(\cdot, \cdot)$ is applied to MEF.

\subsection{Additional Details of FAMO}
Considering $M$ fusion tasks associated with $M$ objectives $\ell_m\}_{m=1}^M$. In $t$-th iteration, the combination weights are obtained by $Z_t=Softmax(\xi_t)$, where $\xi_t\in\mathbb R^{M}$ are unconstrained logits. FAMO updates the model parameters as:
\begin{equation}
    \theta_{t+1}=\theta_t-\alpha\sum_{m=1}^M{\left(C_t\frac{Z_{m,t}}{\ell_{m,t}}\right)\nabla\ell_{m,t}},
\end{equation}
where $C_t=\left(\sum_{m=1}^M{Z_{m,t}/\ell_{m,t}}\right)^{-1}$, $\alpha$ is the step size. And the weighting logits can be updated as:
\begin{equation}
\begin{aligned}
    &\xi_{t+1}=\xi_t-\beta(\delta_t+\gamma\xi_t),\\
    &\delta_t=\begin{bmatrix}\nabla^\top Z_{1,t}\\\vdots\\\nabla^\top Z_{M,t}\end{bmatrix}^\top\begin{bmatrix}\log{\ell_{1,t}}-\log{\ell_{1,t+1}}\\\vdots\\\log{\ell_{M,t}}-\log{\ell_{M,t+1}}\end{bmatrix}
\end{aligned}
\end{equation}
where $\beta$ is the step size, $\gamma$ is the decay. By maximizing the minimum improvement rate, FAMO effectively allocates computational resources and aligns optimization objectives, ultimately improving overall performance. For detailed derivation, please refer FAMO~\cite{famo}.

\subsection{Token Exchange}

TE module~\cite{tokenfusion} operates on the principle that when a uninformative token is detected, it can be replaced with a binary modal token at the corresponding position, preserving essential information while reducing noise:
\begin{equation}
    \begin{aligned}
        X_{i,1}=X_{i,1}\odot \mathbb I_{\phi_{\theta_s}(X_{i,1})\ge \gamma}+X_{i,2}\odot \mathbb I_{\phi_{\theta_s}(X_{i,1})< \gamma},\\
        X_{i,2}=X_{i,2}\odot \mathbb I_{\phi_{\theta_s}(X_{i,2})\ge \gamma}+X_{i,1}\odot \mathbb I_{\phi_{\theta_s}(X_{i,2})< \gamma},
    \end{aligned}
\end{equation}
where $\mathbb I$ is an indicator, and the threshold $\gamma$ is set to $0.02$ according to the paper.

\section{More Results and Analysis}

\begin{table}[]
\centering
\small
\caption{The ablation study for MEF task.}
\begin{tabular}{@{}lll|llll@{}}
\toprule
TI         & TA         & MO         & MI              & FMI             & Qabf            & VIF             \\ \midrule
\multicolumn{3}{c|}{Baseline}        & \textbf{6.6732} & 0.8953          & 0.6748          & 1.4342          \\
\multicolumn{3}{c|}{Baseline-TS}     & 5.2338          & 0.8976          & 0.7154          & 1.3061          \\
\checkmark &            &            & 5.2125          & 0.8974          & 0.7148          & 1.3187          \\
           & \checkmark &            & 5.2231          & 0.8984          & 0.7227          & 1.3154          \\
           &            & \checkmark & 5.9231          & 0.8998          & 0.7039          & 1.4994          \\
\checkmark & \checkmark &            & 5.9976          & 0.8992          & 0.7057          & 1.5153          \\
\checkmark &            & \checkmark & 5.1837          & 0.8989          & 0.7215          & 1.3146          \\
           & \checkmark & \checkmark & 6.1557          & \textbf{0.9002} & \textbf{0.7235} & 1.5274          \\
\multicolumn{3}{c|}{Ours}            & 6.2073          & 0.9000          & 0.7227          & \textbf{1.5338} \\ \bottomrule
\end{tabular}
\label{tab:ablation_mef}
\end{table}

\begin{table}[]
\centering
\small
\caption{The ablation study for MFF task.}
\begin{tabular}{@{}lll|llll@{}}
\toprule
TI         & TA         & MO         & MI              & FMI             & Qabf            & VIF             \\ \midrule
\multicolumn{3}{c|}{Baseline}        & 6.2336          & 0.8777          & 0.5768          & 1.6171          \\
\multicolumn{3}{c|}{Baseline-TS}     & 6.2303          & 0.8822          & 0.6455          & 1.5997          \\
\checkmark &            &            & 6.2566          & 0.8821          & 0.6554          & 1.5931          \\
           & \checkmark &            & 6.2687          & 0.8824          & 0.6834          & 1.5963          \\
           &            & \checkmark & 6.2822          & 0.8833          & 0.6519          & 1.6223          \\
\checkmark & \checkmark &            & 6.3071          & 0.8836          & 0.6523          & 1.6210          \\
\checkmark &            & \checkmark & 6.3229          & 0.8826          & 0.6916          & 1.6050          \\
           & \checkmark & \checkmark & 6.4689          & 0.8841          & 0.6915          & 1.6294          \\
\multicolumn{3}{c|}{Ours}            & \textbf{6.5463} & \textbf{0.8847} & \textbf{0.6973} & \textbf{1.6371} \\ \bottomrule
\end{tabular}
\label{tab:ablation_mff}
\end{table}

\subsection{Ablation Study for MEF and MFF Tasks}

We present the ablation study results for MEF and MFF tasks in \cref{tab:ablation_mef} and \cref{tab:ablation_mff}. The findings on the MFF task align with those of the IVF task, reinforcing the same conclusions. The relatively lower impact of TI on the MEF task may be attributed to its higher reliance on global transformations, where the global operation in TA plays a more significant role.

\begin{table}[]
\centering
\small
\caption{The ablation study on task-invariant integration for MEF task.}
\begin{tabular}{@{}l|llll@{}}
\toprule
     & MI              & FMI             & Qabf            & VIF             \\ \midrule
SF   & 5.2338          & 0.8976          & 0.7154          & 1.3061          \\
IrSF & \textbf{5.2551} & 0.8975          & 0.7125          & 1.3059          \\
IeSF & 5.2512          & \textbf{0.8977} & \textbf{0.7163} & \textbf{1.3074} \\ \midrule
TE   & \textbf{5.2662} & 0.8969          & 0.7096          & \textbf{1.3155} \\
PA   & 5.2428          & 0.8975          & 0.7098          & 1.3056          \\
IPA  & 5.2273          & \textbf{0.8977} & 0.7116          & 1.3048          \\ \bottomrule
\end{tabular}
\label{tab:ablation_ti_mef}
\end{table}

\begin{table}[]
\centering
\small
\caption{The ablation study on task-invariant integration for MFF task.}
\begin{tabular}{@{}l|llll@{}}
\toprule
     & MI              & FMI             & Qabf            & VIF             \\ \midrule
SF   & 6.2303          & 0.8822          & 0.6455          & 1.5997          \\
IrSF & 6.1816          & 0.8818          & 0.6415          & 1.5916          \\
IeSF & \textbf{6.2373} & \textbf{0.8822} & \textbf{0.6516} & \textbf{1.6019} \\ \midrule
TE   & 6.2386          & 0.8817          & 0.6467          & 1.5952          \\
PA   & \textbf{6.2835} & \textbf{0.8825} & 0.6461          & 1.6088          \\
IPA  & 6.2690          & 0.8824          & \textbf{0.6541} & \textbf{1.6091} \\ \bottomrule
\end{tabular}
\label{tab:ablation_ti_mff}
\end{table}

\subsection{Analysis on Task-invariant Integration for MEF and MFF tasks}

We conduct ablation studies to evaluate the effectiveness of IeSF and IPA in Task-invariant Integration for MEF and MFF tasks. As shown in \cref{tab:ablation_ti_mef} and \cref{tab:ablation_ti_mff}, by replacing the IeSF with IrSF or replacing IPA with TE, we observe drops in overall performance, demonstrating the necessity of both IeSF and IPA modules.

\begin{table}[]
\centering
\small
\caption{The ablation study on task-specific adaptation for MEF task.}
\begin{tabular}{@{}l|llll@{}}
\toprule
        & MI              & FMI             & Qabf            & VIF             \\ \midrule
W/o HPF & 6.0651          & 0.8999          & \textbf{0.7247} & 1.4957          \\
W/o ADD & 5.8250          & 0.8995          & 0.7215          & 1.4760          \\
W/o MUL & 6.1305          & 0.8984          & 0.7099          & 1.5164          \\
W/o DW  & 4.8277          & 0.8987          & 0.6894          & 1.2815          \\
Ours    & \textbf{6.2073} & \textbf{0.9000} & 0.7227          & \textbf{1.5338} \\ \bottomrule
\end{tabular}
\label{tab:ablation_ta_mef}
\end{table}

\begin{table}[]
\centering
\small
\caption{The ablation study on task-specific adaptation for MFF task.}
\begin{tabular}{@{}l|llll@{}}
\toprule
        & MI              & FMI             & Qabf            & VIF             \\ \midrule
W/o HPF & 6.2898          & 0.8833          & 0.6801          & 1.6039          \\
W/o ADD & 6.3395          & 0.8845          & 0.6831          & \textbf{1.6385} \\
W/o MUL & 6.3512          & 0.8831          & 0.6663          & 1.6148          \\
W/o DW  & 6.2616          & 0.8840          & 0.6862          & 1.6245          \\
Ours    & \textbf{6.5463} & \textbf{0.8847} & \textbf{0.6973} & 1.6371          \\ \bottomrule
\end{tabular}
\label{tab:ablation_ta_mff}
\end{table}

\subsection{Analysis on Task-specific Adaptation for MEF and MFF tasks}

We conduct ablation studies to evaluate the effectiveness of three branches in Task-adaptive Adaptation for MEF and MFF tasks. As shown in \cref{tab:ablation_ta_mef} and \cref{tab:ablation_ta_mff}, while all branches contribute to the overall performance improvement, their contributions differ. For example, the MFF task relies more heavily on the HPF branch, as clarity is directly related to high-frequency details.

\begin{table}[]
\centering
\small
\caption{The visualization of the dynamic weights on OAF block.}
\begin{tabular}{@{}lll|lll@{}}
\toprule
HPF          & ADD         & MUL        & HPF          & ADD         & MUL        \\ \midrule
\multicolumn{3}{c|}{visible}      & \multicolumn{3}{c}{infrared}      \\
0.0291     & 0.2456    & 0.2054   & 0.0243     & 0.2226    & 0.2729   \\ \midrule
\multicolumn{3}{c|}{over-exposed} & \multicolumn{3}{c}{under-exposed} \\
0.0286     & 0.1963    & 0.3054   & 0.0313     & 0.2575    & 0.1809   \\ \midrule
\multicolumn{3}{c|}{far-focused}  & \multicolumn{3}{c}{near-focused}  \\
0.0159     & 0.2480    & 0.2323   & 0.0133     & 0.2231    & 0.2675   \\ \bottomrule
\end{tabular}
\label{tab:dweight}
\end{table}

\subsection{Visualization of TA}

The weights for each branch of OAF for three fusion tasks is visualized in \cref{tab:dweight}. It can be observed that the weight assigned to the HPF branch is relatively small, as high-frequency details comprise only a minor portion of the overall information. Nevertheless, as demonstrated in the ablation study on task-specific adaptation, these high-frequency components play a critical role in boosting performance.

\end{document}